# Intelligent Hotel ROS-based Service Robot

Yanyu Zhang, Xiu Wang, Xuan Wu, Wenjing Zhang, Meiqian Jiang and Mahmood Al-Khassaweneh

*Abstract*—With the advances of artificial intelligence (AI) technology, many studies and work have been carried out on how robots could replace human labor. Based on the conventional surveys, this paper presents a conceive on the intelligence hotel robot, which simplify the check-in process. Also considered in different environment settings, robot can self-adaption with the different cases. Combined with Hokuyo Lidar and Kinect Xbox camera, the robot can plan the routes accurately, besides it can reach to different floors. Focus on the analysis of the numbers on the elevator's wall, we assume that robot has the ability to read the numbers and symbols, then press the corresponding button. What is more, the intelligent voice system provides an assistant for the customers. Furthermore, the intelligent order sending system provides fully-automatic food delivery. However, the hotel not only divides human work, but also reconstructs it from tasks. Moreover, the purpose of reconstruction is not simply for replacement of works. Such modification of a task is often observed taking place in human- system interactions. It is an extremely productive process of labor emerging in this area.

**Keywords -- Arm manipulating, deep learning, logical control, robotic object detection, sensors apply.**

## I. INTRODUCTION

The birth of modern robots marks the beginning of the realization of the human dream of intelligent machines [1]. At the present stage, grassroots work in the hotel industry is dull, because of the intensity of simple and repeated errand running tasks, coupled with the traditional concept of "the hotel is to serve" constraints. It is difficult to attract a new generation of labor, [2] coupled with the current wage level has been unable to let the hotel to recruit ideal staff. Therefore, in order to reduce the labor costs, reliable replacement is needed. According to the "2017 annual hotel turnover survey report" released by MTA, 20%-40% employees leave the hotel industry each year.

According to media reports, some customers have stayed an extra night in the hotel in order to experience the special hotel robot. The emergence of hotel robot is an application of the artificial intelligence. It is a new technical science to research and develop theories, methods, technologies and application systems for the simulation, extension and expansion of human intelligence.

Nowadays, with the development of the IoT (Internet of things) [3], various types of robots are applied to carry out the transportation and guiding mission in the various environments like hotels [4], medical centers [5] and shopping malls [6]. To transport to different floors, the robot can analyze the destination and plan the nearest route. [7] With the support of Hokuyo Lidar and Kinect camera, it is possible to understand the surrounding environment.

The conceive of hotel robots have 4 functions. (i) When the guests arrive the setting region, the intelligent voice system will greet to them. Also voice system is used to remind guests in the following sections [8]. (ii) After check in, the G-mapping manager can plan the nearest route to the destination depending on the room number typed on the digital keyboard. (iii) Besides, the robot can lead the customer to the corresponding room. In the process of getting to the room, the robot will consider taking the elevator if necessary. Also, mechanical arm needed to press the button in the elevator and pick up food during the automatic food delivery. (iv) In addition, the robot also has transport service. When consumers purchase Hotel's order, robots have the ability to complete the automatic delivery service.

## II. THEORY AND APPROACHES

### A. General Information

Using pioneer-3dx robot combined with Lidar, Kinect camera and speaker (as shown in figure 1), intelligent hotel robot is constructed. We developed the navigation and G-mapping, combining with the MATLAB, ROS and Eclipse.

The general idea of the traveling is divided into four sections: (i) Lead the robot out of the initial room. (ii) Keep the robot in the middle line of the corridor and come into the elevator. (iii) After reaching the necessary floor through elevator, the robot will lead customers to the room. (iv) The robot will return to the initial room.

The route setting is based on D* arithmetic. The algorithm works by selecting a node from the open list iteratively and evaluating it. Then, it propagates the node's changes to all the neighboring nodes and places them on the open list. This propagation process is termed "expansion". In contrast to canonical A* [9], which follows the path from start to finish, D* begins by searching backwards from the goal node. Each expanded node has a back pointer which refers to the next node leading to the target, and each node knows the exact cost of the target. When the start node is the next node to be expanded, the algorithm is done, and the path to the goal can be found by simply following the back pointers.

To improve the intelligent and reduce the human labor at the same time, we provide a method to send orders by robot through the QR code. The main idea is when the customers take an order, the hotel center will send a QR code to the customer. At the same time, the robot will receive the information on the order and take food by mechanical arms in the setting areas and store them in the robot storage. After that robot will bring the goods to the correct room. When the robot arrived, the customer can open the door of the robotic storage through the QR code. The whole process is fully-automatic.

### B. Pioneer 3-DX

Pioneer robots are smaller than most, but they pack an impressive array of intelligent mobile robot capabilities that rival bigger and much more expensive machines. The Pioneer 3-DX with onboard PC is a fully autonomous intelligent mobile robot. With its powerful ARCOS server and advanced MOBILEROBOTS client software, the Pioneer 3 shown in figure 2 is fully capable of mapping its environment, finding its way home and performing other sophisticated path-planning tasks [10].

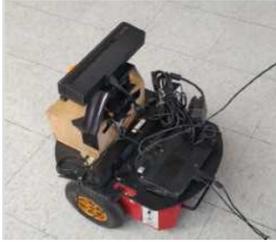 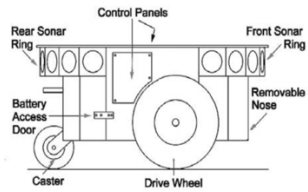

Fig. 1. Pioneer 3-DX   Fig. 2. Pioneer 3-DX Features [11]

For the Pioneer 3-DX, it provides various client-server connection options. In all cases, that client PC must connect to the internal host or user control panel serial port for the robot and software to work. For the piggyback laptop or embedded PC, the serial connection is via a common "pass-through" serial cable. Radio Ethernet is a little more complicated but is the preferred method because it lets you use many different computers on the network to become the robot's client. (Figure 3)

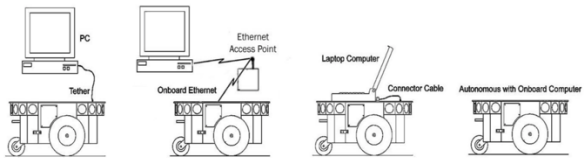

Fig. 3. Client-server connection options [11]

ARCOS-based robots support up to four sonar arrays, each with up to eight transducers that provide object detection and range information for collision avoidance, features recognition, localization, and navigation. The sonar positions in all Pioneer 3 sonar arrays are fixed: one on each side, and six facing outward at 20-degree intervals. (Figure 4) In this way, we can scan the environment and avoid obstacles better.

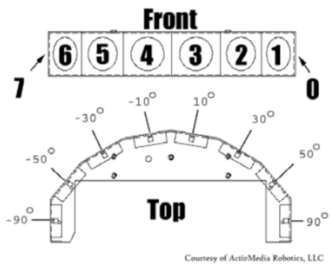

Fig. 4. Sonar Array [11]

C. Lidar

Hotel robots need Lidar (Light Detection and Ranging) [12] to scan and record a defined floor map while avoiding obstacles. It is an on-board laser scanning using GPS (Global Position System) and IMU (Inertial Measurement Unit). The sensor emits a laser beam and travels through the air to the surface of the ground or objects, and then reflects through the surface. The reflected energy is received by the sensor and recorded as an electrical signal.

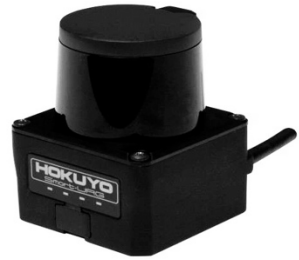

Fig. 5. HOKUYO Lidar

If the time of transmission and the time of reception are accurately recorded, the distance (R) of the laser to the ground or the surface of the object can be calculated by the following formula:

$$R = \frac{ct}{2}$$

Where (c) is speed of light and (t) is difference between the time of transmission and the time of acceptance.

Lidar is used to scan the surrounding environment when the robot avoids obstacles. It can measure the return time of each ray at one moment to get the distance of the obstacle from the robot. When multiple consecutive rays return for a long time or do not return, then there is no obstacle in that direction. Therefore, Lidar mainly scans out the surrounding environment in the function of avoiding obstacles.

D. G-mapping and Navigation

Our fundamental purpose is based on navigation. The theory of navigation is D* arithmetic. The robot plans a reasonable route to the destination through the navigation system. Navigation; a two-dimensional function package in ROS; is simply based on the information flow of the sensor such as the input odometer and the global position of the robot. Then, the safe and reliable robot speed control command is calculated through the navigation algorithm (refer to figure 6). By the path cost of a node comes from the previous node to the goal. In classical grid-based planning this value is computed as

$$g(s) = \min_{s' \in nbrs(s)} (c(s,s') + g(s')),$$

where nbrs (s) is the set of all neighboring nodes of s, c (s, s′) is the cost of traversing the edge between s and s′, and g(s′) is the path cost of node s′.

We plan to travel the route in Global frame: the overall path planning according to a given target destination. In the navigation of ROS, the global route of the robot to the target location is first calculated through global path planning. This feature is implemented by the "navfn" package. Navfn calculates the minimum cost path on the costmap through the Dijkstra optimal path algorithm as the global route of the robot. In the future, the VFH algorithm or others should be added to the algorithm.

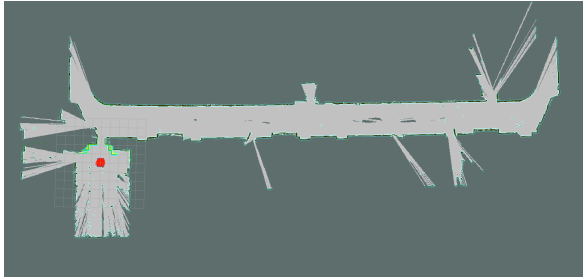

Fig. 6. RVIZ

### E. Speaker and voice library

To make our service look better, we added a speaker to report the current statement while leading customers to the room. We record the voice and add it into the library. The speaker converts an electrical signal into an acoustic signal. The performance of the speaker has a great influence on the sound quality. The speaker is one of the weakest devices in audio equipment, and it is one of the most important components for sound effects.

### F. KINECT XBOX ONE

At the end, we installed a camera on the robot to determine whether the elevator door is open by judging the distance of the robot itself to the target position, which is a good way to increase the accuracy (Figure 7). The digital camera converts the analog video signal generated by the video capture device into a digital signal and stores it in a computer. The digital camera captures the image directly and transmits it to the computer via a serial, parallel or USB interface.

We use the camera for two purposes: to judge the switch on the elevator door and let the information in the camera's field of view be transmitted to the robot. After that, you need to analyze the image, which requires OpenCV software.
OpenCV is a cross-platform computer vision library based on BSD license (open source) distribution that runs on Linux systems. It consists of a series of C functions and a small number of C++ classes. It also provides interfaces to languages such as Python, Ruby, and MATLAB, and implements many general algorithms for image processing and computer vision.

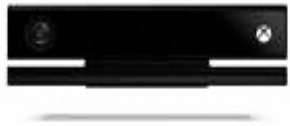

Fig. 7. KINECT XBOX ONE

### III. EXPERIMENTS

We conducted two experiments using Pioneer and HOKUYO Lidar both in the engineering building during the weekdays. (i) The first one is leading the customers to the room within the same floor; (ii) the second is leading them to different floors. In addition, during the running of the robot, there were some people in the corridors, which are considered as obstacles. Taking that in to consideration, we set the experiments during the weekdays and keep the initial point to be in the same position. Our aim is to measure the accuracy and efficiency of the robot to finish the leading. Our data include the points setting in the middle of the journey and the travel time.

The engineering building has three floors as shown in figure 8, 9 and 10. The robot will lead to the room in any floors after you input a room number on the keyboard.

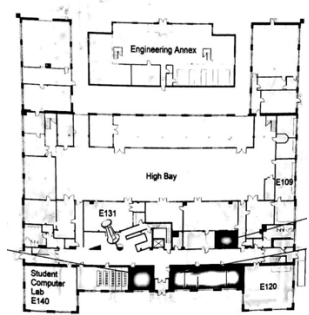 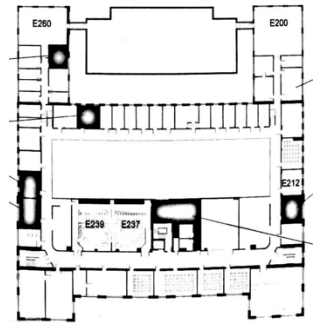

Fig. 8. First floor      Fig. 9. Second floor

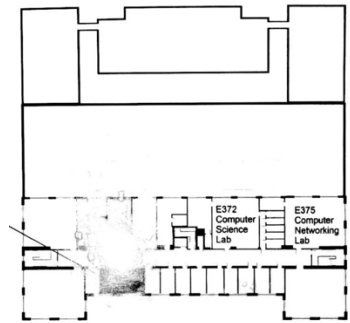

Fig. 10. Third floor

In the first experiment, we set the initial point at the E240 in the engineering building. And we tested three points in the second floor: E234, E236, E238 and E230. In the second experiment, we set two rooms on the third floor: E320, E315. We recorded the traveling time, the success rate and the Lidar information for each time.

### IV. RESULTS

In the first experiment, we assume the destination is on the same floor. So, the robot does not need to lead customers to the elevator. We assume four rooms on the second floor: 238, 236, 234, 230 (refer to figure 11). The points that were set in the middle of the starting and destination are shown in figure 11. We used these points to drive the robot to reach the destination. The robot passes these points one by one and records the time it needs to arrive everyone. We can find that the average success rate is more than 80%. On the other hand, with the increasing of the journey, the accuracy rate dropped dramatically. This can be related to the long distance navigation that brings some errors to the wheel editor. Also, we found that the accuracy rate increases with the increase of the points in the middle of the journey. Furthermore, the traveling time decreases with the increase of the points at most of time as shown in figure 12.

| Room Number | Times | Point 1 | Point 2 | Point 3 | Point 4 | Point 5 | Point 6 | Point 7 |
|---|---|---|---|---|---|---|---|---|
| 238 | Travel 1 | (4.58,-0.48,0) | (3.27,-4.64,0) | (4.56,-4.56,0) | (4.29,-0.0458,0) | (0,0,0) | / | / |
|  | Travel 2 | (4.58,-0.48,0) | (4.56,-4.56,0) | (3.27,-4.64,0) | (4.56,-4.56,0) | (4.29,-0.0458,0) | (0,0,0) | / |
|  | Travel 3 | (4.58,-0.48,0) | (4.29,-0.046,0) | (4.56,-4.56,0) | (3.27,-4.64,0) | (4.56,-4.56,0) | (4.29,-0.0458,0) | (0,0,0) |
| 236 | Travel 1 | (4.58,-0.48,0) | (4.4,-10.8,0) | (3.34,-10.7,0) | (4.29,-0.0458,0) | (0,0,0) | / | / |
|  | Travel 2 | (4.58,-0.48,0) | (4.4,-10.8,0) | (3.34,10.7,0) | (4.4,-10.8,0) | (4.9,0.00614,0) | (0,0,0) | / |
|  | Travel 3 | (4.58,-0.48,0) | (4.89,-5.48,0) | (4.4,-10.8,0) | (3.34,-10.7,0) | (4.4,-10.8,0) | (4.9,0.00614,0) | (0,0,0) |
| 234 | Travel 1 | (4.58,-0.48,0) | (3.54,17.1,0) | (4.79,-17.3,0) | (4.29,-0.0458,0) | (0,0,0) | / | / |
|  | Travel 2 | (4.58,-0.48,0) | (4.79,-13.3,0) | (3.54,-17.1,0) | (4.79,-17.3,0) | (4.29,-0.0458,0) | (0,0,0) | / |
|  | Travel 3 | (4.58,-0.48,0) | (5.15,-7.56,0) | (4.79,-17.3,0) | (3.54,-17.1,0) | (4.79,-17.3,0) | (4.29,-0.0458,0) | (0,0,0) |
| 230 | Travel 1 | (4.58,-0.48,0) | (4.74,-23,0) | (5.14,-23.4,0) | (4.29,-0.0458,0) | (0,0,0) | / | / |
|  | Travel 2 | (4.58,-0.48,0) | (5.14,-23.4,0) | (4.74,-23,0) | (5.14,-23.4,0) | (4.29,-0.0458,0) | (0,0,0) | / |
|  | Travel 3 | (4.58,-0.48,0) | (5.27,-13.7,0) | (5.14,-23.4,0) | (4.74,-23,0) | (5.14,-23.4,0) | (4.29,-0.0458,0) | (0,0,0) |

Fig. 11. Second Floor Test Point Data

| Room Number | Times | Time 1 | Time 2 | Time 3 | Time 4 |
|---|---|---|---|---|---|
| 238 | Travel 1 | 01:52.0 | 02:31.0 | 02:58.0 | 02:47.0 |
|  | Travel 2 | 01:55.0 | 02:11.0 | 02:08.6 | 02:11.1 |
|  | Travel 3 | 01:58.7 | 02:06.3 | 02:44.0 | 02:24.7 |
| 236 | Travel 1 | 03:41.2 | 02:59.7 | 03:31.3 | 03:18.5 |
|  | Travel 2 | 02:51.5 | 02:58.6 | 03:34.3 | 03:33.1 |
|  | Travel 3 | 03:12.3 | 03:27.7 | 02:49.9 | 03:04.6 |
| 234 | Travel 1 | 03:25.2 | 03:59.6 | 04:08.7 | 03:48.6 |
|  | Travel 2 | 04:10.5 | 04:08.4 | 03:56.3 | 03:53.9 |
|  | Travel 3 | 04:06.6 | 03:55.7 | 03:59.3 | 04:01.7 |
| 230 | Travel 1 | 04:46.5 | 05:13.5 | 05:08.6 | 05:05.9 |
|  | Travel 2 | 05:10.5 | 05:02.4 | 04:58.4 | 05:12.6 |
|  | Travel 3 | 05:09.6 | 04:49.6 | 05:13.7 | 04:57.6 |

Fig. 12. Second Floor Traveling Time Data

In the second experiment, we set the destination to the third floor. We found that the robot is away from the setting point with the increase of the route. The practical position of the robot has a 0.5-1 meter gap as the value of the odometry. As shown in figure 13 and 14, these two tests are starting at the same initial point and the surrounding environments are nearly the same. But the first test is obvious away from the setting point in front of the elevator. However, the robot also can travel to the correct destinations on the third floor.

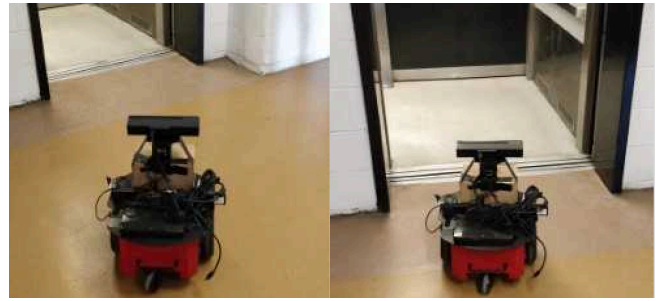

Fig. 13. Elevator Test 1     Fig. 14. Elevator Test 2

## V. DISCUSSION

During the test process, we found that the error on the wheel editor have a big influence whit the increase of the total journey. So, we give two methods to deal with this problem. (i) First, we unify the sensors to scan the map and navigation. Because we are using Pioneer 3dx sonar arrays for scanning the map of the Engineering and Lidar for leading the way, there exit error during flipping stage between the sensors. (ii) Second, use additional sensors to coordinate with the Lidar can decrease the error. Double-loop can rise the accuracy.

The data from the RVIZ basically meet the set value. In some cases, the robot will have a slope facing the door, which will lead to the robot scan the environment by rotating. This will spend a large amount of time.

## VI. CONCLUSION AND FUTURE WORK

In this study, we tested the robot's ability of leading toward to the different cases. We chose the different floors as the simulation in the real social space. And we provide an imagination that the robot can complete the automatic delivery service.

In the future, service robots are people-oriented, so they must be a combination of intelligent and humanized. So, we provide three aspects that need to improve: (i) It is necessary to expand the D* arithmetic. Because base on the D* arithmetic, sometimes robot cannot find the way in a short time. We plan to try the A* and VFH arithmetic [13] to find the best way. (ii) Besides, combine with the robotic arm to open the elevator or press the correct floor number inside the elevator are needed. (iii) To improve the intelligent and at the same time reduce the human labor, we provide a method to send orders by robot through the QR code.

Essentially, a robot is a mobile computer connected to hotel applications (such as PMS). Therefore, it is easy for the robot to obtain the relevant information of the guest and provide the corresponding personalized service. The robot is a computer, so it can handle the languages of different countries, and it can process data anytime and anywhere, which will improve the guest check-in experience.

## VII. ACKNOWLEDGEMENTS